\documentclass{article}

\usepackage{arxiv}

\usepackage[utf8]{inputenc} % allow utf-8 input
\usepackage[T1]{fontenc}    % use 8-bit T1 fonts
\usepackage{hyperref}       % hyperlinks
\usepackage{url}            % simple URL typesetting
\usepackage{booktabs}       % professional-quality tables
\usepackage{amsfonts}       % blackboard math symbols
\usepackage{nicefrac}       % compact symbols for 1/2, etc.
\usepackage{microtype}      % microtypography
\usepackage{lipsum}
\usepackage{graphicx}
\usepackage{graphicx}
\usepackage{algpseudocode}
\usepackage{algorithm}
\usepackage{amsmath}

\title{Knowledge Distillation with Training Wheels}

\author{
 Guanlin Liu \\
   \And
 Anand Ramachandran \\
  \And
 Tanmay Gangwani \\
  \And
 Yan Fu \\
    \And
 Abhinav Sethy \\
\And
 Amazon Alexa AI \\
 yanfu@amazon.com
}

\DeclareMathOperator*{\argmax}{arg\,max}

\begin{document}
\maketitle
\begin{abstract}
Knowledge distillation is used, in generative language modeling, to train a smaller student model using the help of a larger teacher model, resulting in improved capabilities for the student model. In this paper, we formulate a more general framework for knowledge distillation where the student learns from the teacher during training, and also learns to ask for the teacher's help at test-time following rules specifying test-time restrictions. Towards this, we first formulate knowledge distillation as an entropy-regularized value optimization problem. Adopting Path Consistency Learning to solve this, leads to a new knowledge distillation algorithm using on-policy and off-policy demonstrations. We extend this using constrained reinforcement learning to a framework that incorporates the use of the teacher model as a test-time reference, within constraints. In this situation, akin to a human learner, the model needs to learn not only the learning material, but also the relative difficulty of different sections to prioritize for seeking teacher help. We examine the efficacy of our method through experiments in translation and summarization tasks, observing trends in accuracy and teacher use, noting that our approach unlocks operating points not available to the popular Speculative Decoding approach.
\end{abstract}

% keywords can be removed

\section{Introduction}

While Large Language Models (LLM) with parameter sizes in the tens or hundreds of billion \cite{dettmers2022gpt3} \cite{metallama3} have shown excellent language, reasoning, and knowledge capabilities, they also come with increased inference costs. It is true that such generalist frontier models excel in a wide-range of tasks, but smaller models may suffice to satisfy the operational needs for a subset of tasks.

Knowledge Distillation (KD) \cite{hinton2015distilling} has been used as a means to improving the capabilities of smaller models. Typically, in KD, a smaller student model is used to mimic a larger teacher model, over some training data corresponding to tasks of interest. The training data can be a curated training set, can be elicited from the teacher model for the tasks of interest, or can be "on-policy" data generated from the student model.

Typically, the distilled student models are deployed stand-alone at test time \cite{agarwal2024policy} \cite{gu2023minillm}.  As student models are much smaller than teacher models, the improvement in the quality of results through distillation has a hard ceiling. In this paper, we rework the knowledge distillation formulation to allow the student model to use teacher help at test time based on testing rules expressed as natural language instructions. Pedagogically, this teaches the student new auxiliary lessons. The student learns the relative difficulty of different parts of the curriculum in a fine-grained manner and uses it to seek help as efficiently as possible. Practically, this allows the student to overcome the hard limitations of its size and accept expert help, when needed, at test time, resulting in a decoding method that allows a flexible trade-off between accuracy and efficiency.

There exist algorithms in the speculative decoding family \cite{chen2023accelerating}\cite{leviathan2023fast}, where a larger target model is used to validate tokens generated by a smaller draft model, and rejection sampling is employed to replace draft tokens with target tokens where appropriate. This results in improvement in latency without loss of accuracy. Knowledge distillation has been studied as a way to improve speculative decoding efficiency \cite{DistllSpec} \cite{liu2023online}, and speculative decoding has been employed in a lossy fashion where not all rejected draft tokens are resampled from the target \cite{DistllSpec}, which allows a trade-off between accuracy and efficiency.  The hallmark of all of these approaches is the use of target (teacher) model to judge and determine which tokens from the draft (student) to keep and which to reject. In these cases, the student is not aware of the teacher at runtime, and the teacher is deployed for each token generated at least in a validation capacity.  This deviates from our method where the student directly determines teacher use as appropriate without teacher involvement. The speculative decoding family of algorithms, thus, miss the pedagogical aspect of our work, which supplements the normal KD objectives with auxiliary learning functions. In addition, the use of the teacher model to validate tokens, and the use of rejection sampling to discard several tokens from the draft model, results in lower efficiency compared to ours.

% This means, the speculative decoding approaches miss the pedagogical aspect of our work, and al

In our approach, to incorporate teacher use at test time, we first examine the KD objective, specifically, the reverse KL-divergence (KLD) formulation of it, which has several advantages over the forward KLD approach \cite{huszar2015not}.  Prior work \cite{gu2023minillm} determined that reverse KLD formulation is equivalent to policy gradient optimization. However, the authors noted the need for several techniques such as single-step decomposition, length normalization, and teacher mixin to stabilize training and improve reward hacking. In this paper, we derive the reverse KLD objective as an instance of value optimization instead, and solve it using Path Consistency Learning (PCL) \cite{DBLP:journals/corr/NachumNXS17}.  The use of PCL allows us to natively utilize off-policy demonstrations from the teacher model, supplementing on-policy ones, to obtain the necessary training stability and convergence without any additional techniques. We further extend this value optimization framework within a constrained Reinforcement Learning formulation to incorporate the use of the teacher model during test time within constraints expressed via Natural Language prompts.

We empirically observe the behavior of models trained using our approach for two tasks - translation and summarization. For each task, a single model is trained to observe several rules regarding teacher use, from no teacher use to significant teacher use during test time. We observe that the models are able to learn to adhere to teacher use rules with high fidelity, and observe how they prioritize tokens to self-generate and tokens for which to request teacher help. We also observe how output quality changes with allowed teacher use.

In the next section, we introduce our methodology including formulation and training approaches. In the section after that, we outline our experiments and results.

\section{Methodology}

\subsection{Knowledge Distillation is Value Optimization under Entropy Regularization}
\label{sec:KD_derive}

Let $\log \Pi(X) = \sum_i \log \pi(x_i|X_{<i})$
represent the log-probability of a sequence,
$X$, parameterized by the student model. Similarly, let
$\log P(X) = \sum_i \log p(x_i|X_{<i})$
represent the log-probability of a sequence, $X$, parameterized by the teacher model.
We use the short-hand notation
$E_{X_{i:T} \sim \Pi(.|X_{<i})} \log \frac{\Pi(X_{t:T}|X_{<t})}{P(X_{t:T}|X_{<t})} = E_{i:T}$. 
Note that the sequence prefix, $X_{<i}$, can include the prefix of a generated sequence as well as some task-specific instruction.

The following result holds for all autoregressive models, including Large Language Models.

\begin{equation}
    E_{i:T} = E_{x_i \sim \pi(.|X_{<i})} \Big(\log \frac{\pi(x_i|X_{<i})}{p(x_i|X_{<i})} + E_{i + 1:T}\Big)
    \label{eq:exp_recur}
\end{equation}

Setting, $-E_{i:T} = V^{\pi}(X_{<i})$, we get the following rewrite, with constants $\tau=1,\gamma=1$.

\begin{align}
    V^{\pi}(X_{<i}) = &\sum_{x_i} \pi(x_i|X_{<i})\Big(\log p(x_i|X_{<i}) - \nonumber \\
    &\tau \log \pi(x_i|X_{<i}) + \gamma V^{\pi}(X_{<i + 1})\Big) \label{eq:kd_recur}
\end{align}

Equation \ref{eq:kd_recur} represents the Bellman equations for an entropy-regularized Markov Decision Process with deterministic dynamics and state represented by $s_i = X_{<i}$.  The optimal policy for this formulation obeys certain path consistency constraints over trajectories sampled from on or off-policy rollouts \cite{DBLP:journals/corr/NachumNXS17}. This means that reverse KLD knowledge distillation can be optimized through minimizing the PCL loss over on- and off-policy demonstrations.

As mentioned before, prior work has shown that reverse KLD can be optimized through policy gradient optimization \cite{gu2023minillm}. There are a few differences that may be noted through our alternative formulation as entropy-regularized value optimization. First, in our formulation, the KL-term is decomposed into a reward term, $\log p(x_i|X_{<i})$ and an entropy regularization term, $\log \pi(x_i|X_{<i})$. This leads to a different interpretation of Knowledge Distillation as maximizing the likelihood of student-generated sequences under the teacher, with a regularization term encouraging exploration. Specifically, the role of one of the KLD terms, namely, $\log \pi(x_i|X_{<i})$, is to encourage exploration during Monte Carlo sampling of on-policy rollouts. If we are mainly interested in the student model learning the primary modes of the teacher, we can deboost this term using the $\tau$ component in order to improve training stability, if we have a large number of sequence prefixes $X_{<i}$ to start rollouts from. In language modeling, such prefixes may be obtained from training data or task instructions.

Second, in policy-gradient optimization, several techniques need to be used to stabilize training such as mixing in teacher tokens and weighting them by importance weights and so forth. In PCL, path consistency must be obeyed for trajectories from off-policy distributions as well without any modifications to the loss. Hence, this offers us a natural means to incorporate teacher demonstrations in the training loop, without any changes to the formulation.

In practice, for language modeling tasks, we are given sequence prefixes, and our goal is to generate the suffix optimally. To explicitly express this, conditioning on $X_{<i}$ can be rewritten as conditioning on $X_{<i},Y$, where $Y$ is a task instruction that was never part of the generation process, but which nevertheless forms part of the sequence prefix. We may then rewrite all conditioning on $X_{<i}$ in Equations \ref{eq:exp_recur}, \ref{eq:kd_recur} as conditioning on $(X_{<i},Y)$, where $Y \sim \mathcal{D}$, a dataset of task instructions. Hence, our optimization goal becomes,

\begin{eqnarray*}
    \pi^* &=& \argmax_{\pi} E_{Y \sim \mathcal{D}} V^{\pi}(X_{<1}, Y) \\
    &=& \argmax_{\pi} E_{Y \sim \mathcal{D}} V^{\pi}(Y)
\end{eqnarray*}

\subsubsection{Proof of Equation \ref{eq:exp_recur}}

Let $\log \Pi(X) = \sum_i \log \pi(x_i|X_{<i})$ represent the log-probability of a sequence $X$ parameterized by the 
student model. Similarly, let $\log p(X) = \sum_i \log p(x_i|X_{<i})$ represent the log-probability of a sequence
parameterized by the teacher model. We start with the following factorization for language models
(and autoregressive models in general):

\begin{align}
    &E_{X_{i: T} \sim \Pi(.|X_{<i})} \log \frac{\Pi(X_{i: T}|X_{<i})}{P(X_{i: T}|X_{<i})} = \label{eq:kd_recur1} \\
    &E_{x_i \sim \pi(.|X_{<i})} E_{X_{i + 1:T \sim \Pi(.|X_{<i+1}})} \log \frac{\Pi(X_{i: T}|X_{<i})}{P(X_{i: T}|X_{<i})} = \label{eq:kd_recur2} \\
    &E_{x_i \sim \pi(.|X_{<i})} \Big(\log \frac{\pi(x_i|X_{<i})}{P(x_i|X_{<i})} + \nonumber\\
    &E_{X_{i + 1:T \sim \Pi(.|X_{<i+1}})} \log \frac{\Pi(X_{i + 1: T}|X_{<i + 1})}{P(X_{i + 1: T}|X_{<i + 1})}\Big)
\end{align}

Designating $E_{X_{i: T} \sim \Pi(.|X_{<i})} \log \frac{\Pi(X_{i: T}|X_{<i})}{P(X_{i: T}|X_{<i})}$ as $E_{i:T}$, we get

\begin{equation}
E_{i:T} = E_{x_i \sim \pi(.|X_{<i})} \Big(\log \frac{\pi(x_i|X_{<i})}{p(x_i|X_{<i})} + E_{i+1:T}\Big)
\end{equation}

which is the same as Equation \ref{eq:exp_recur}

\subsection{Incorporating test-time Teacher use into Knowledge Distillation}
\label{sec:ConstrainedKD}

In the prior section, we derived a method to do KD by formulating it as a Reinforcement Learning value optimization problem whose goal is to maximize the likelihood of student-generated sequences as evaluated by the teacher.

In this section, we examine methods to let the student call for teacher help at test time. That is, we want to expand the action space of the student to include a call for teacher help. This can be implemented by simply letting the student generate a special token indicating the teacher call. However, without constraints, the student model would call the teacher to generate every token, rendering the method useless. It is useful only if the student model is constrained to call the teacher for help a limited number of times. This naturally leads to a constrained reinforcement-learning formulation, where the primary objective is maximization of output log-likelihood as evaluated by the teacher, and the secondary objective is a constraint related to teacher use. We further expand this idea to allow several operational settings (teacher use fractions) in a single model, all specified through the sequence prefix $Y$.

To derive this, we start by noting that PCL optimizes a policy, $\pi$ that maximizes the value function $V^{\pi}$ shown in Equation \ref{eq:kd_recur}. We add a second value function to this corresponding to the teacher use constraint, $V_C$. The optimization objective becomes

\begin{equation}
    \pi^* = \argmax_{\pi} V^{\pi}(Y)\;s.t.\;V_C(Y) \leq 0 \label{eq:optim}
\end{equation}

We consider $V_C$ of the form

\begin{align}
V_C(X_{<i}, Y) = &\sum_{x_i} \pi(x_i|X_{<i}, Y)(r_C(X_{<i},x_i,Y) + \nonumber \\
&\gamma V_C(X_{<i+1}, Y))
\end{align}

Using Lagrange relaxation, we may write a new value function.

\begin{align}
    V_C^{\pi}(X_{<i}, Y) &= V^{\pi}(X_{<i}, Y) - \lambda V_C(X_{<i}, Y) \nonumber \\
    &\sum_{x_i} \pi(x_i|X_{<i}, Y)\Big(\log p(x_i|X_{<i},Y) \nonumber \\
    &- \lambda r_C(X_{\leq i}, Y) \nonumber \\
    &- \tau \log \pi(x_i|X_{<i},Y) \nonumber \\
    &+ \gamma V_C^{\pi}(X_{\leq i, Y})\Big) \label{eq:constrained_pcl}
\end{align}

The constrained optimization problem of Equation \ref{eq:optim} can then be solved as follows, following prior practice in constrained Reinforcement Learning \cite{tessler2018reward} \cite{borkar2005actor}.

\begin{equation}
    \pi^* = \min_{\lambda \geq 0} \max_{\pi} V_C^{\pi}(Y)
\end{equation}

We implement this practically using primal and dual updates as follows.
\begin{itemize}
    \item Update $\pi$ by minimizing Path Consistency Loss
    \item Updating $\lambda = \Gamma_{\lambda}\Big(\lambda - \eta_{\lambda} \nabla_{\lambda} V_C^{\pi}(Y) \Big)$, where $\Gamma(x)$ maps $R \rightarrow R^+$ monotonically. $\nabla_{\lambda} V_C^{\pi}(Y) = -V_C(Y)$
\end{itemize}

\subsection{Student-Teacher decoding and data representation}

The student model's action space is the vocabulary of the language being modeled plus a special action to seek help from the teacher. The teacher-seeking action is invoked by the special token, $<\tau>$, which results in a teacher token being sampled.

It is true that for all student tokens, when an action is sampled, $x_i$, the next state deterministically becomes $X_{\leq i}$. However, when the token $<\tau>$ is sampled from the student, the value of the next state depends on the teacher token being sampled. Note that the teacher, here, is part of the environment and not part of the policy. Hence, when token $<\tau>$ is sampled from the student, the next state is non-determinstic, when teacher sampling is not greedy. This is not reflected in our treatment above, as the equations implicitly express deterministic dynamics, for the sake of brevity. However, PCL works equally well for non-deterministic dynamics and hence any method of sampling from the teacher model is supported during training.

In the sequel, the following notation is used: $C_{<i} = (X_{<i}, Y)$. Let $x_{i}^{(s)}$ represent an output token from the student at time step $i$, and $x_{i}^{(t)}$ represent an output token from the teacher model at time step $i$. We have the following relations.

\begin{equation}
x_{i}^{(s)} \sim \pi(x|C^{(S)}_{<i}) \label{eq:stud_decode}
\end{equation}

\begin{equation}
x_{i}^{(t)} \sim p(x|C^{(T)}_{<i}) \label{eq:teacher_decode}
\end{equation}

Note that, $x_{i}^{(s)} \in \mathcal{V} \cup \{<\tau>\}$ and $x_{i}^{(t)} \in \mathcal{V}$, where $\mathcal{V}$ is the token alphabet of the language being modeled. $C^{(S)}$ and $C^{(T)}$ represent sequence prefixes for the student and teacher, respectively.

We define two sequences, $X^{(out)}$ and $X^{(in)}$ as follows.

\begin{equation}
    x^{(in)}_i =
    \begin{cases}
    x_i^{(s)}, &\text{if } x^{s}_i \neq <\tau> \\
    x_i^{(t)} + ||\mathcal{V}||, &\text{otherwise}
    \end{cases} \label{eq:X_in}
\end{equation}

\begin{equation}
    x^{(out)}_i =
    \begin{cases}
    x_i^{(s)}, &\text{if } x^{s}_i \neq <\tau> \\
    x_i^{(t)}, &\text{otherwise}
    \end{cases} \label{eq:X_out}
\end{equation}

$X^{(out)}$ represents the output of the decoding algorithm, and $X^{(in)}_{<i}$ represents the state of the decoding process at time $i$. Note that $X^{(in)}$ includes additional internal information regarding the decoding process that is useful to the student model to budget teacher-use.  That is, $C^{(S)}_{<i} = (X^{(in)}_{<i}, Y)$ and $C^{(T)}_{<i} = (X^{(out)}_{<i}, Y)$. 

As $X^{(in)}$ is processed by the student model, each token in $X^{(in)}$ is embedded by it. For tokens in the alphabet, $\mathcal{V}$, we use the default embedding scheme in the student model, and for the special token, $<\tau>$, we add a learnable embedding to the default embedding.

\begin{equation}
    e(x^{(in)}_i) = \begin{cases}
        Embed(x^{(in)}_i),&\text{if } x^{(in)}_i \in \mathcal{V} \\
        Embed(x^{(t)}_i) + \beta,&\text{otherwise}
    \end{cases} \label{eq:embedding}
\end{equation}

\subsection{Training Scheme}

We train the student model in two phases. The first phase initializes the model to the action space, and the second stage finetunes it to use the actions appropriately, within budget.

\subsubsection{Phase 1 Training}

In the first phase, we train the student through behavior cloning to help it learn the basics of how to use the new special token, $<\tau>$, and the implications of using it.  For behavior cloning, we define an oracle policy that we teach the student through teacher-forcing. The oracle policy is not realizable in practice.

Given a teacher trajectory, we calculate the KL-divergence between the student and teacher distributions at each position along the trajectory and rank the sequence positions according to their KL-divergence in decreasing order. Let the KL-divergence rank of position $i$ be indicated as $r_i$.  Given a trajectory length, $l$ and a teacher use budget, $b$, we obtain a subset of positions, $S_{hi}$ which contains the top $l \times b$ positions by rank, $r_i$. Let $KL_i = D_{KL}(p(.|C^{(S)}_{<i})||\pi(.|C^{(T)}_{<i}))$, where the student distribution is renormalized to tokens in $\mathcal{V}$.

\begin{equation}
    loss = \begin{cases}
        -\log \pi(<\tau>|C^{(S)}_{<i}), &\text{if } i \notin S_{hi} \\
        KL_i,&\text{otherwise}
    \end{cases}
\end{equation}

This scheme teaches the student to use the teacher model at positions where the KL-divergence is high and improve its own distribution to align with the teacher's where the KL-divergence is low.

\subsubsection{Phase 2 Training}

Phase 1 training initializes the model teaching it to use the special $<\tau>$ token.  However, Phase 1 is not on-policy, as the model learns based on teacher trajectories. Since the student model is of substantially lower capacity than the teacher model, the trajectories produced from it are unlikely to match those from the teacher model. This can cause downstream modeling fidelity issues such as due to exposure bias \cite{ranzato2015sequence}, and the student use may not strictly adhere to constraints on teacher use budget as it is a constrained optimization problem that is not modeled correctly when using supervised learning.

We continue training using Reinforcement Learning through PCL as outlined above. The reward for the model is as follows, where $b$ is the teacher use budget and $\lambda$ is the Lagrangian. The teacher use budget represents what fraction of tokens should arise from the teacher on average.

\begin{equation}
    r_i = \begin{cases}
        \log p(x^{(s)}_i|X_{<i}) + \lambda b,&\text{if }x^{(s)}_i \neq <\tau> \\
        \lambda(b - 1), &\text{otherwise} \label{eq:rewards}
    \end{cases}
\end{equation}

This modifies the formulation in Equation \ref{eq:constrained_pcl} slightly, by using the teacher likelihood rewards only for tokens generated by the student. That is, we approximate the reward assuming that teacher tokens are implicitly correct.

In addition, there is a reward bonus $\lambda b$ when the token is generated by the student. When the token arises from the teacher, there is a penalty, $\lambda(b - 1)$.  Note that the teacher-use reward terms cancel out when the student exactly matches the budgeted teacher use, is positive when it satisfies the constraint strictly and is negative otherwise. This corresponds to the constraint value function given below ($1_{()}$ is the indicator function).

\begin{align}
    V_C(X_{<i, Y}) = E_{x_i \sim \pi}\Big(1_{x^{(s)}_i = <\tau>} \nonumber \\
    - 1_{x^{(s)}_i \neq <\tau>} b + \gamma V_C(X_{<i+1}, Y)\Big)
\end{align}

By tuning $\lambda$ during training we ensure that $\lambda$ moves towards zero if the constraints are met. Hence when constraints are met for a long training period, the reward for student token generation is $\log p(x^{(s)}_i|X_{<i})$ and is $0$ for teacher tokens (teacher tokens are implicitly trusted).

We use two other modifications in practice for ease and stability: (1) the reward component, $r = \log p(x|X_{<i})$ is converted to $\frac{r - \mu_r}{\sigma_r}$, where $\mu_r$ is the mean of teacher token likelihoods, and $\sigma_r$ is the standard deviation of the student token likelihoods (as evaluated by the teacher). In addition, the total reward is clipped to $[-2, 2]$. (2) We base $\lambda$ on a proxy of $V_C(Y)$, $O_C(Y)$, which simply calculates $\mathbb{E}(b - \sum_{i=1}^{l}1_{x^{(s)}_i = <\tau>})$. In other words, $\lambda$ is updated based on the excess teacher use compared to the teacher use budget. This is estimated from the most recent rollouts.

In our experiments, we train a single model to abide by several teacher use budgets, $b$.  To do this, we insert a decoding instruction into the model's task instruction. For instance, when the student should not use any teacher participation, the task instruction "summarize: " is changed to "With no teacher use, summarize: ", and for the corresponding rollouts, we set $b=0$.

\section{Results}

\subsection{Methods compared}

Whitebox distillation, like the one presented in this paper, where the student has complete access to the teacher's output distribution, has been studied before in the context of lossy speculative decoding with student and teacher models as draft and target models in DistillSpec \cite{DistllSpec}. In this paper, we use DistillSpec as the baseline for comparison.

In speculative decoding, a teacher model is used to verify, in parallel, multiple tokens sampled from a student model. The number of tokens sampled at a time from the student is referred to as the draft length. If verification fails, the token for which it fails is sampled from a new distribution combining the student and teacher distributions. Parallel verification allows the teacher to be used efficiently. In standard speculative decoding, the verification criterion is stringent ensuring that the speculative decoding output (or distribution) is the same as that of the teacher model. In lossy speculative decoding (i.e., DistilSpec), the student token is accepted under more relaxed conditions allowing for trading off latency of decoding with accuracy. This is done via a lenience factor that relaxes the rejection criterion by a multiplicative fraction between 0 and 1. Details on our implementation of lossy speculative decoding are in the Appendix. Following the original work on Knowledge Distillation for Speculative decoding, we use Generalized Knowledge Distillation (GKD) \cite{agarwal2023gkd} to distil the teacher model onto the student model.

Our method doesn't involve speculation or verification. In our case, all decisions on token generation or teacher use are made by the student model, and the choice is not contested by the teacher. Hence every student token generated is accepted as an output token. Due to this lack of student token wastage, we believe our method will unlock additional performance gains.
% While it is reasonable to expect that the absence of verification may impact accuracy due to the inability to evaluate tokens at test , we expect that the fact that our own Knowledge Distillation step is performed in a manner accommodating teacher use, will mitigate these effects to a large degree, with significant performance gains.

\subsection{Models and Data}

For our experiments we used FLAN-T5-SMALL as the student model and FLAN-T5-XL as the teacher model \cite{chung2024scaling}. We performed experiments using translation for which we used WMT-19 German to English translation dataset \cite{barrault2019findings}, and summarization using the XSum dataset \cite{xsum-emnlp}.

For our method, for each task we trained the student model to obey six different teacher-use budgets, which would be communicated to the model at test time through the input prompt. The budgets are $b \in \{0.0, 0.1, 0.2, 0.3, 0.4, 0.5\}$ referred to by the keywords, "no", "light", "moderate-light", "moderate", "high" and "very high" teacher use. For a specific budget, the corresponding keyword would be included in the prompt. E.g., if we expect the model to use the teacher model, on average, less than 10\% of the time, we would prompt it as, "With light teacher use, summarize: ".

For lossy speculative decoding, we train the student model following the GKD approach, where both student and teacher trajectories are used to minimize token-wise reverse KL-divergence of the student's output distribution. For inference using lossy speculative decoding, we vary the lenience factor from 0 to 1 in steps of 0.1.

\subsection{Training Details}

For our approach, we performed 200 batches of phase 1 training through behavioral cloning, and a subsequent 5000 batches of RL finetuning following the reward scheme in Equation \ref{eq:rewards}. During RL finetuning, two trajectories are sampled for each task, one an off-policy trajectory according to the oracle policy, and one an on-policy trajectory. We performed four gradient updates per batch of episode generation, with a data mix that contains samples from a replay buffer, as well as oracle trajectories and on-policy trajectories that were generated in the current batch. After training is completed, we select the checkpoint that satisfies teacher-use constraints for all teacher-use budgets within a tolerance value $\delta$ and maximizes the output quality (measured through ROUGE-2), on a randomly sampled subset of 512 examples from the validation set.
% For testing, for each bucket, $b$, we selected the best checkpoint corresponding to that bucket. The best checkpoint for a bucket $b$ is defined as a checkpoint which achieves a target use fraction in the range $[b - \delta, b + \delta]$ ($\delta$ is a small tolerable error value), with the best output quality over 512 examples sampled from the validation set.

For DistillSpec, we trained the student model using a mix of teacher and student trajectories, following GKD. For each batch of trajectories sampled, we performed four gradient updates. Similar to our approach, we trained GKD over 5200 batches of trajectory sampling (or 20800 gradient updates). For testing, we selected the checkpoint with the best output quality over 512 examples sampled from the validation set.

\subsection{Performance Metrics}

For both the summarization and translation tasks, we compare the decoding efficiency, in terms of latency and output quality represented using ROUGE-2. For DistilSpec, we used three draft lengths, 3, 5, and 10. We use greedy sampling (temperature = 0) for all experiments. All experiments are performed on a single NVIDIA T4 GPU.

When a draft cycle is perfectly accurate (every token passes verification), a token may be sampled from the teacher model for free. Thus for a draft-length of 3, at best, the teacher model will be used 25\% of the output tokens. However, the effective number of times the teacher is used can be higher in real life, as draft tokens may be discarded. We used a draft length of 10 to correspond to a lower bound of teacher usage fraction of 1 / 11, which is lower than the lowest non-zero target-use configuration for our approach (which is 1/10).

\begin{figure}[h!]
    \centering
    \includegraphics[width=\columnwidth]{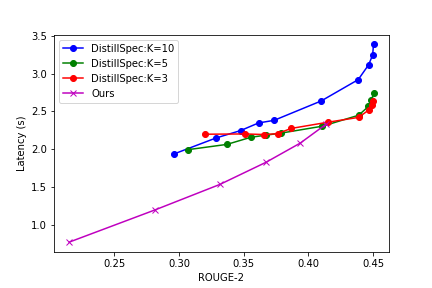}
    \caption{Performance metrics for the translation task. The X-axis indicates output quality. The Y-axis represents the achieved latency}
    \label{fig:translation}
\end{figure}

The performance metrics for the translation task are provided in Figure \ref{fig:translation}. In the figure, the most desirable operating points are in the lower right corner. Notably, all approaches show a linear trend in latency-vs-quality trade-off for lower latency points, and show a saturating behavior at higher latency. This is very encouraging as it shows that operating points with higher efficiency can capture outsized quality improvements from the teacher model. DistillSpec configurations for draft lengths 3 and 5 operate similar to each other. However, for draft length 10, which has the lowest possible theoretical teacher usage, the achieved latency is, higher in most cases. The operating characteristics of DistillSpec and our approach are shifted with respect to each other. Our approach operates in the lower latency regime capable of making more aggressive trade-offs than DistillSpec. From a latency perspective, our approach is strictly better than DistillSpec. In areas where the operating points overlap in terms of quality, our approach provides lower latency than DistillSpec by up to approximately 25\%.

\begin{figure}[h!]
    \centering
    \includegraphics[width=\columnwidth]{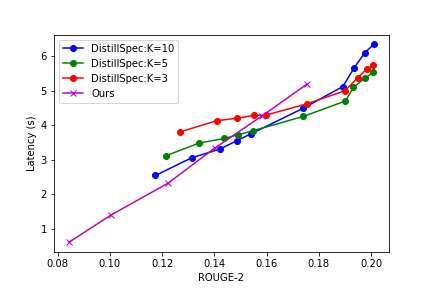}
    \caption{Performance metrics for the summariation task. The X-axis indicates output quality. The Y-axis represents the achieved latency}
    \label{fig:summarization}
\end{figure}

The performance metrics for the summarization task are provided in Figure \ref{fig:summarization}. Similar to the translation task, our approach unlocks operating points not available in DistilSpec, specifically at lower latencies. At higher latencies, DistilSpec is more than competitive with our approach, with similar or slightly better latency-accuracy trade-offs.

Across both tasks, our approach shows an ability to deliver operating points between that of the student and teacher model operating characteristics . Our approach represents a more aggressive trade-off between latency and accuracy than DistillSpec, and it is able to explore operating regions not accessible via DistillSpec.

\section{Conclusions}

In this paper, we presented a formulation of Knowledge Distillation as a value optimization problem. This formulation has some practical advantages brought forth by the ability to seamlessly use off-policy and on-policy trajectories for model optimization. We then extended this method by expanding the action space of the model to include a functionality to call the teacher model for help during test time, within given teacher use budget. This was derived to be a constrained reinforcement learning extension to the knowledge distillation formulation. We trained models on two tasks, translation and summarization. For each task we defined six different rules corresponding to six teacher use budgets at test time.  We found that the models are largely able to satisfy the provided constraints, and show desirable improvement in output quality with allowed teacher usage. This approach does not depend on generating tokens based on validation or rejection sampling. Therefore, unlike prior speculative decoding approaches with knowledge distillation, our approach can extract larger efficiency gains without appreciable loss of accuracy.

We believe our framework can be extended to multi-agent decoding setups with a light-weight student model acting as the arbitrator between multiple sources of expertise. These sources can be other models, or tools, each with specific accuracy and latency/throughput characteristics.

\section{Appendix}

\subsection{Operating points}

The operating points for Knowledge Distillation With Training Wheels are shown in the table below.

\begin{table}[h!]
    \centering
    \caption{Prompt instructions for teacher use and corresponding usage fraction targeted}
    \begin{tabular}{cc}
        \hline \\
        \textbf{Teacher Use Instruction} & \textbf{Fraction} \\
        \hline \\
        With no teacher use, \{summarize|translate\} & 0.0\\
        With light teacher use, \{summarize|translate\} & 0.1\\
        With moderate-light teacher use, \{summarize|translate\} & 0.2\\
        With moderate teacher use, \{summarize|translate\} & 0.3\\
        With high teacher use, \{summarize|translate\} & 0.4\\
        With very high teacher use, \{summarize|translate\} & 0.5\\
        \hline\\
    \end{tabular}
    \label{tab:my_label}
\end{table}

\subsection{Training Details}

For both approaches we used a learning rate of $10^{-5}$ and weight decay of $10^{-5}$ for the policy (student model). For our approach, we used a learning rate of $10^{-6}$ and weight decay of $10^{-6}$ for the value network. For constrained RL in our approach, we used a Lagrangian learning rate of $1e-2$. The Lagrangian is capped at $2$. The Lagrangian is learnt for each operating point separately, and its gradients are calculated based on teacher use in several trailing batches. For our approach, we turned off dropout in both policy and value networks.

For both GKD and phase 2 training in our approach, for each task in the training set, we sampled one trajectory from the student model and one trajectory from the teacher model. For GKD we used reverse KL Divergence as the loss for the sampled trajectories.

\subsection{Lossy greedy speculative decoding}

Lossy speculative decoding was first presented in the original speculative decoding paper \cite{leviathan2023fast} and later studied in the context of Knowledge Distillation in DistillSpec \cite{DistllSpec}. Here, we present the algorithm used in this paper, specifically for the greedy sampling setting.

\begin{algorithm}[h!]
\caption{Single step of lossy speculative decoding: Require teacher, $p$, student, $\pi$, context, $(X_{<i}, Y)$, lenience, $l$, draft length, $K$}
\begin{algorithmic}[1]
    \For{$j$ from $0$ to $K - 1$}
        \State $x_{i+j} \gets \argmax_{x} \pi(x|X_{<i + j}, Y)$
        \State $\hat{x}_{i + j} \gets \argmax_{x} p(x|X_{<i + j}, Y)$
        \State $X_{i + j: i + j + 1} \gets [x_{i + j}]$
    \EndFor
    \State $n \gets \min \Big(\{j | p(x_{i + j}|X_{<i + j}, Y) < l p(\hat{x}_{i + j}|X_{<i + j}, Y), \forall\;0 \leq j < K, j \in N\} \cup \{K\}\Big)$
    \If{$n = K$}
    \State $X_{i + K} \gets \argmax_{x} p(x|X_{<i + K - 1}, Y)$
    \Else
    \State $X_{i + n: i + n + 1} \gets [\hat{x}_{i + n}]$
    \EndIf
    \State Return {$X_{i:i + n}$}
\end{algorithmic}
\end{algorithm}

\bibliographystyle{splncs04}
\bibliography{mybib}

\if0
\bibliographystyle{unsrt}  


\fi

\end{document}